\documentclass[conference]{IEEEtran}
\IEEEoverridecommandlockouts

\usepackage[utf8]{inputenc}
\usepackage{ifthen}
\usepackage{hyperref}

\usepackage[dvipsnames]{xcolor}
\usepackage{soul}
\usepackage[normalem]{ulem}
\usepackage{graphicx}

\usepackage{epsf,picinpar}
\usepackage{varioref}
\usepackage{fdsymbol}

\usepackage{enumitem}

\usepackage[backend=biber,style=ieee,maxcitenames=2,maxbibnames=99,citestyle=numeric-comp]{biblatex}

\usepackage{pifont}

\usepackage{orcidlink}
\usepackage{booktabs}

\usepackage{listings}

\usepackage{multicol}
\usepackage{multirow}
\usepackage{makecell}
\usepackage{caption}
\usepackage{subcaption}
\usepackage[framemethod=TikZ]{mdframed}

\usepackage{dblfloatfix}
\usepackage{nicefrac}

\usepackage[color=red!50!white,textsize=scriptsize]{todonotes}

\usepackage{fontawesome5}

\usepackage{makecell}

\usepackage{url}
\newcommand{\secref}[1]{Sec.~\ref{#1}}
\newcommand{\figref}[1]{Fig.~\ref{#1}}
\newcommand{\tabref}[1]{Tab.~\ref{#1}}

\newcommand{\assignedto}[1]{%
    \ifthenelse{\boolean{showannotations}}%
    {\textbf{\noindent\ding{46}\textcolor{white}{~\colorbox{\assignementcolor}{Assigned to:}}~\textcolor{\assignementcolor}{#1}\\}%
    }
    {}
}

\renewcommand{\nb}[4]{
    {\fcolorbox{gray}{#2}{\bfseries\sffamily\scriptsize{#1}}
	{\sf\small$\blacktriangleright$\textcolor{#4}{\textit{#3}}$\blacktriangleleft$}}%
}

\newcommand\personmarker[3]{\noindent\nb{#1}{yellow}{#2}{#3}}

\newcommand\sr[1]{\noindent\personmarker{Stefan}{#1}{blue}}

\newcommand{\rem}[1]{%
    \ifthenelse{\boolean{showannotations}}%
    {\textcolor{\oldtextcolor}{\st{#1}}}%
    {}%
}

\newcommand\add[1]{%
    \ifthenelse{\boolean{showannotations}}%
    {\textcolor{\newtextcolor}{{#1}}}%
    {#1}%
}

\newcommand\addblockbegin{%
    \ifthenelse{\boolean{showannotations}}%
    {\color{\newtextcolor}}%
    {}%
}

\newcommand\addblockend{%
    \ifthenelse{\boolean{showannotations}}%
    {\color{black}}%
    {}%
}

\newcommand\rep[2]{%
    \ifthenelse{\boolean{showannotations}}%
    {\rem{#1}~\add{#2}}%
    {#2}%
}

\makeatletter
 \if@todonotes@disabled
 
 \else
 
 \fi
 \makeatother

\makeatletter
 \if@todonotes@disabled
 \newcommand{\donemarker}[2][green]{#2}
 \else
 \newcommand{\donemarker}[2][green]{%
 {\ifthenelse{\equal{#2}{}}{\todo[color={#1}]{DONE UNTIL HERE.}}%
 {\todo[color={#1}]{#2}}}%
 }
 \fi
 \makeatother

\newcounter{tempOcounter}
\stepcounter{tempOcounter}
\newcounter{tempRcounter}
\stepcounter{tempRcounter}
\newboolean{showannotations}
\setboolean{showannotations}{true} 

\hypersetup{pdfborder={0 0 0}}

\newcommand{\newtextcolor}{blue}
\newcommand{\oldtextcolor}{red}
\newcommand{\assignementcolor}{orange}
\definecolor{highlightcolor}{rgb}{.99, 1, .0}
\sethlcolor{highlightcolor}
\definecolor{orcidlogocol}{HTML}{A6CE39}

\mdfsetup{%
   backgroundcolor=gray!5,
   middlelinewidth=1pt,
   roundcorner=7pt}

\title{Engineering Automotive Digital Twins on Standardized Architectures: A Case Study}

\author{
     \IEEEauthorblockN{Stefan Ramdhan\orcidlink{0009-0002-6976-0043}\textsuperscript{\faCanadianMapleLeaf},
     Winnie Trandinh\orcidlink{0009-0005-9882-5557}\textsuperscript{\faCanadianMapleLeaf},
     Istvan David\orcidlink{0000-0002-4870-8433},
     Vera Pantelic\orcidlink{0000-0003-1696-2768},
     Mark Lawford\orcidlink{0000-0003-3161-2176}}
     \IEEEauthorblockA{McMaster Centre for Software Certification (McSCert), Hamilton, Canada}
     \IEEEauthorblockA{McMaster University, Hamilton, Canada}
}

\begin{document}

\maketitle
\begingroup\renewcommand\thefootnote{\faCanadianMapleLeaf}
\footnotetext{S. Ramdhan and W. Trandinh contributed to the paper equally.}
\endgroup


\begin{abstract}
Digital twin (DT) technology has become of interest in the automotive industry. There is a growing need for smarter services that utilize the unique capabilities of DTs, ranging from computer-aided remote control to cloud-based fleet coordination.
Developing such services starts with the software architecture. However, the scarcity of DT architectural guidelines poses a challenge for engineering automotive DTs.
Currently, the only DT architectural standard is the one defined in ISO 23247. Though not developed for automotive systems, it is one of the few feasible starting points for automotive DTs.
In this work, we investigate the suitability of the ISO 23247 reference architecture for developing automotive DTs.
Through the case study of developing an Adaptive Cruise Control DT for a 1/10\textsuperscript{th}-scale autonomous vehicle, we identify some strengths and limitations of the reference architecture and begin distilling future directions for researchers, practitioners, and standard developers.
\end{abstract}

\section{Introduction}\label{sec:intro}

Digital twins (DT) are real-time and high-fidelity computational reflections~\cite{maes1987concepts} of real assets~\cite{tao2019digital}, offering cost-efficient and safe alternatives for interacting with the asset and controlling it~\cite{kritzinger2018define}.
Digital twinning has been a transformative paradigm in the engineering of complex cyber-physical systems~\cite{gomes2024foundational}, many of which are safety-critical, such as manufacturing systems~\cite{yang2025live}, power grids~\cite{zhou2019digital}, and biophysical systems~\cite{david2023digital}. The success of DTs in such safety-critical domains made the case for DTs in the automotive industry too~\cite{piromalis2022digital}, marked by an increased interest in automotive DTs~\cite{deng2023systematic}.

The complexity of DTs warrants careful architectural design. Unfortunately, the relative scarcity of architectural standards and reference architectures (RAs) for DTs hinder their design, development, and by extension, their adoption. This is a particularly acute problem in heavily standardized domains, such as automotive.
Currently, the only DT-specific architectural standard is the relatively new ISO 23247 -- Digital Twin Manufacturing Framework~\cite{iso23247}. Published in 2021 by the International Organization for Standardization (ISO), the standard defines a development framework for digital twins, particularly in the manufacturing domain. Part 2 of the standard provides a detailed, layered RA designed and developed for manufacturing systems.
The other few available DT standards are rather high-level and conceptual, and do not define a RA---e.g.,
ISO/IEC 30173:2023 (Digital twin — Concepts and terminology)~\cite{iso30173}.
As a consequence, DTs outside manufacturing are currently architected through standards that offer solutions to sub-problems of DT engineering.
%
%
Such limitations have been reported in an array of domains, e.g., industrial IoT~\cite{vukovic2021digital} and systems of systems~\cite{adesanya2024systems}.
In the absence of better-suited alternatives, the  ISO 23247 RA has drawn increasing interest, e.g., in edge computing~\cite{kang2025edge} and AI simulation~\cite{liu2025ai}, and its adaptation for some domains has already begun, e.g., in aerospace~\cite{shtofenmakher2024adaptation} and battery systems~\cite{cederbladh2024towards}.


In this paper, we investigate the fitness of the ISO 23247 RA for automotive DTs. Through a case study of a DT for the adaptive cruise control (ACC) system in a 1/10\textsuperscript{th}-scale autonomous vehicle, we identify benefits and limitations of the RA, and open a discussion about standardized automotive DT architectures.
%
To facilitate reuse, we make our research artifacts (source code and documentation) publicly available.\footnote{\url{https://github.com/McSCert/Automotive-Digital-Twin}} In addition, a demonstration of the DT is publicly available.\footnote{\url{https://youtu.be/Rv-sJJQc3WU}}

\section{Background and related work}\label{sec:background}

\subsection{The ISO 23247 standard and reference architecture}

In 2021, the International Standards Organization (ISO) published the \textit{ISO 23247 -- Digital twin framework for manufacturing} standard~\cite{iso23247} that defines a development framework for DTs. The standard is organized into four parts:
1) Overview and general principles,
2) Reference architecture,
3) Digital representation of physical manufacturing elements, and
4) Information exchange.
For our purposes, Part 2 is the most relevant as it defines an entity-based RA, with each entity is responsible for implementing a part of the DT's functionality.

The \textit{Observable Manufacturing Element (OME)} is any item that has an observable physical presence or operation in manufacturing, e.g., equipment, facilities, products, personnel, processes, etc. 
The \textit{Data Collection and Device Control Entity} is responsible for collecting data from, and controlling the OME.
The \textit{Digital Twin Entity} digitally represents the OME.
Finally, the \textit{User Entity} provides interfaces for users to interact with the \textit{Digital Twin Entity}, e.g., monitoring data or controlling the OME.  
Each entity is composed of \textit{Functional Entities} (FEs), which are smaller units responsible for a particular task such as actuating the OME, transforming data into a human-interpretable format, or interfacing with the user. The standard does not give guidelines as to how FEs are to be implemented. For example, one can choose to use OPC Unified Architecture (UA) of IEC 62541 to facilitate the implementation of the Data Translation FE, or implement a custom communication layer. 
    


Although the standard was developed for the manufacturing domain, its RA is an appealing feature for DT developers in other domains.
%
\textcite{shtofenmakher2024adaptation} adapt the RA of ISO 23247 to space debris detection. In their approach, the monolithic User Interface FE is decomposed into four distinct FEs, each providing a specific viewpoint for space data systems, public users, and satellite owners and operators, who need to interact with the DT in very different ways. Since not all space objects are controllable, the OME is decomposed into controllable and uncontrollable Resident Space Objects. Both types of OMEs are able to provide data through the Data Collection and Control Entity.
\textcite{cederbladh2024towards} investigate the fitness of the RA for DTs of heterogeneous battery systems in electric vehicles. In their proposal, the DT Entity is split into two components to execute on separate hardware. A lightweight real-time component deals with time-sensitive computation and an offline component deals with computation-intensive tasks. 
\textcite{liu2025ai} investigate the fitness of the RA for AI simulation, the technique of training AI agents through simulated data. The proliferation of DTs has had a positive impact on simulator technology too: today, some of the most advanced simulators are found in DTs, making an appealing case for implementing AI simulation by DTs. In addition, DTs can collect data from the physical twin through purposeful experimentation~\cite{mittal2023towards}. The \textit{Collection Identification FE} is an apt spot to implement such features, while AI agents can integrate with the DT through the User Entity.

These works underscore the utility of the ISO 23247 RA outside manufacturing. Of course, part of this utility is due to the lack of alternatives. Nonetheless, by properly mapping domain-specific requirements, the RA can serve as a useful starting point in the development of DTs outside of manufacturing.
Our work is the first attempt at investigating the fitness of the ISO 23247 RA for automotive systems.

\subsection{Automotive digital twins}
In the automotive context, the \textit{ego vehicle} refers to the vehicle of interest hosting hardware and software that enable autonomous driving. Streams of data from LiDAR, cameras, and other sensors are fused in real time to detect objects in the ego vehicle's environment. Each object is typically assigned an ID and is characterized by its state estimate---e.g., size, position, velocity---which is updated every perception cycle, and is called an \textit{object track}.
Using these tracks, ACC can be developed, which is an advanced driver assistance system (ADAS) that adjusts the ego vehicle's speed to maintain a safe following distance behind the vehicle in its driving path, called the \textit{lead vehicle}. The driver can set a \textit{desired speed} to be maintained when no lead vehicle is present. 


DTs can also be used to streamline the development of vehicles and ADAS systems, or improve their operation~\cite{hossain2023applications}.
During the design phase of automotive systems, DTs are often used as cyber-physical experimentation facilities, thanks to their advanced simulator components and the tight coupling between the virtual and physical environments~\cite{hu2024how}.
This is often achieved in combination with 3D virtual technology, e.g., 3D CAD design~\cite{shikata2019digital}. 
DTs are also used to validate compliance with requirements~\cite{son2021digital}.
During vehicle operation, DTs are used for a variety of optimization and health monitoring purposes.
\textcite{ramtilak2005digital} use a DT of spark ignition to improve fuel economy and driving performance.
\textcite{magargle2017simulation-based} use DTs to collect vehicle data and analyze vehicle performance to inform drivers.
\textcite{rajesh2019digital} implement predictive maintenance of a brake system using a DT, integrating a state-of-the-art simulator, CREO Simulate, into it.
\textcite{venkatesan2019health} implement a DT for health monitoring and prognosis of a permanent magnet synchronous motor.
DTs are also used as a product lifecycle management tool in support of digital threads of vehicles, establishing a continuous, end-to-end information flow throughout the lifecycle~\cite{heber2017towards}.
Additional insights into the role and applications of DTs in automotive are available in the surveys of \textcite{deng2023systematic} and \textcite{schwarz2022applications}.

\section{The Adaptive Cruise Control Case Study}\label{sec:case-background}

The goal of our work is to investigate the fitness of the ISO 23247 RA~\cite{iso23247} for developing automotive DTs.
To achieve this goal, we choose the case study empirical genre, which is ``\textit{an empirical inquiry that investigates a contemporary phenomenon (the case) in depth and within its real-world context, especially when the boundaries between phenomenon and context are unclear}''~\cite{yin2018case}. As such, case studies provide particularly detailed and credible evidence of the investigated phenomenon.
Specifically, we conduct an evaluative case study by developing the DT for the ACC of a 1/10\textsuperscript{th}-scale autonomous vehicle, then assess the benefits and limitations of the ISO 23247 architecture in a real-world setting.

\subsection{Context}

DT technology enables an array of benefits for ACC through off-loading any critical or non-critical computation for which resource-constrained electronic control units (ECU) are typically not sufficient.
%
To implement the ACC-DT, we develop the following three services through a DT.

\subsubsection{Data Monitoring} 
Monitoring ACC-relevant data flowing through a vehicle is crucial for tracking system performance. The DT presented in this case will contain a user interface (UI) allowing a user to view, save, and generate reports of vehicle state data. Additionally, the UI will display real-time vehicle state data, the projected driving path, and detected object tracks in the vehicle's proximity.

\begin{figure*}
    \centering
    \begin{subfigure}{.48\textwidth}
        \centering
        \includegraphics[height=4.5cm]{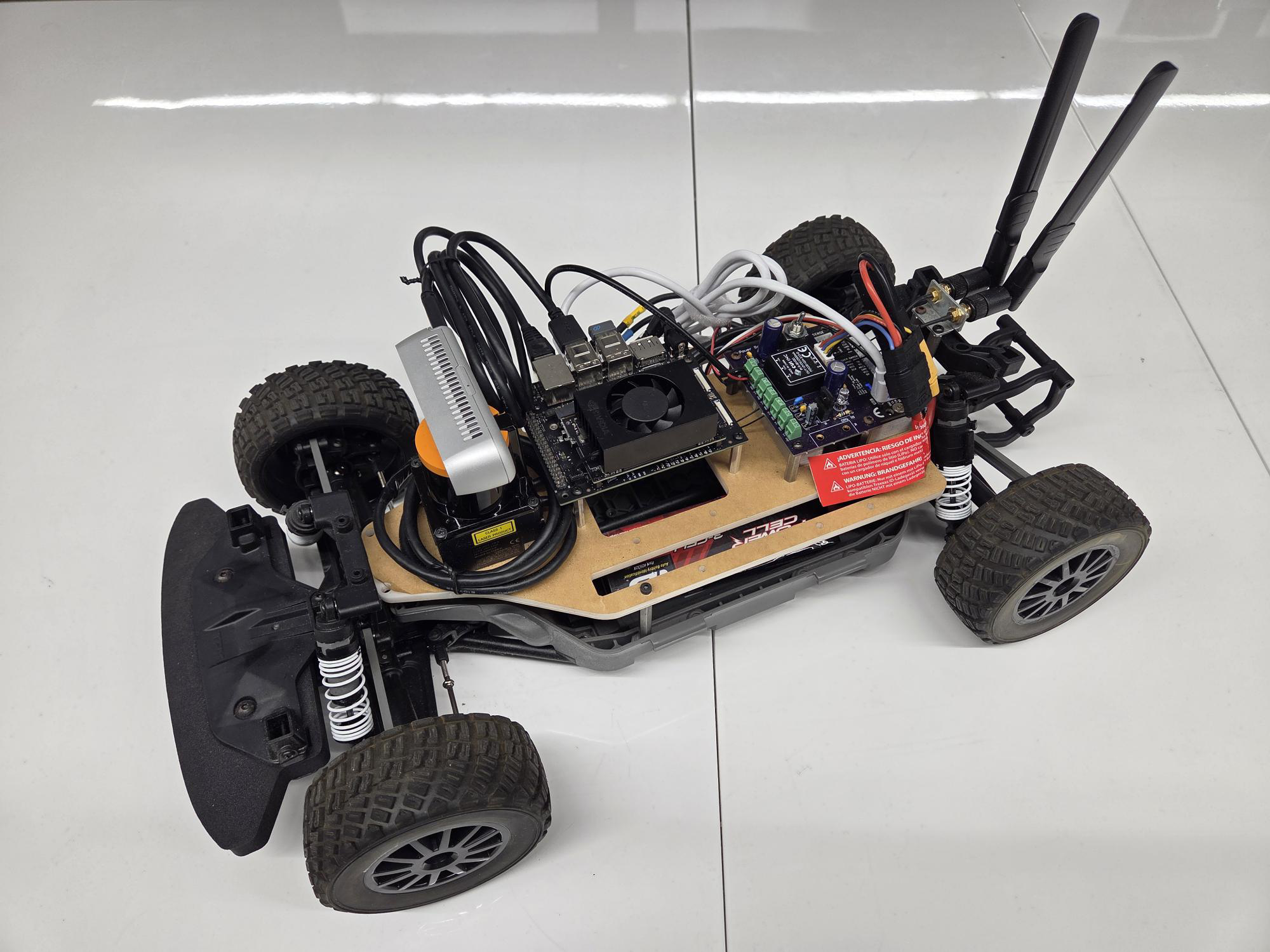}
        \caption{Base hardware: teleoperated vehicle retrofitted with advanced sensor and control technology}
        \label{fig:car}
    \end{subfigure}%
    \hfill%
    \begin{subfigure}{.48\textwidth}
        \centering
        \includegraphics[height=4.5cm]{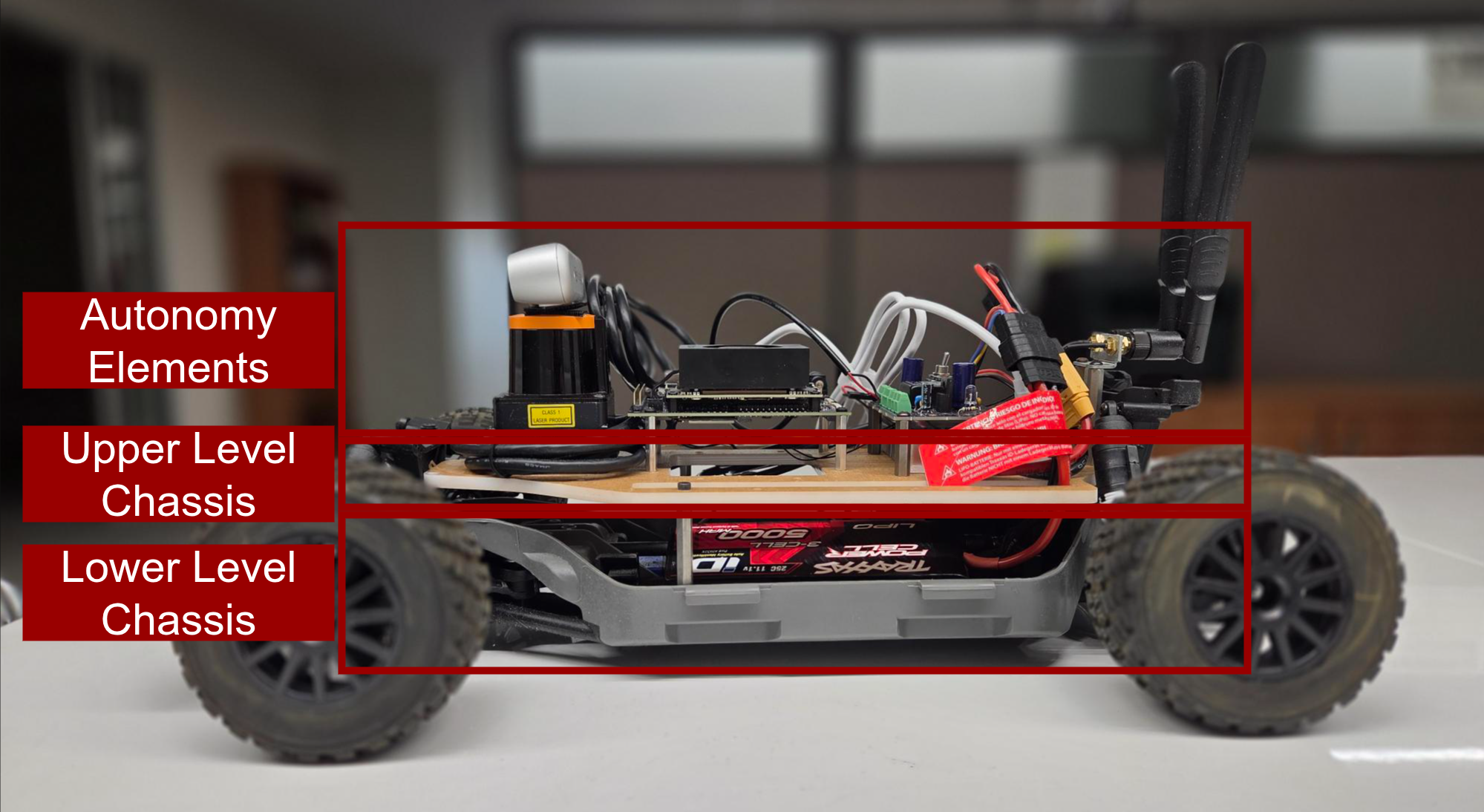}
        \caption{Main hardware layers of the physical twin: lower and upper level chassis, and the autonomy elements}
        \label{fig:hw-arch}
    \end{subfigure}
    \caption{The physical twin: a 1/10\textsuperscript{th}-scale autonomous vehicle}
    \label{fig:pt}
\end{figure*}

\subsubsection{Computation Offloading} 
A DT can alleviate the computational burden placed on resource-constrained on-board hardware by allocating compute to better-suited equipment, e.g., in the cloud. In this case study, we demonstrate this functionality by offloading only the ACC system from the vehicle to the DT. The ACC system determines the lead vehicle, then computes a desired speed based on the distance to the lead vehicle, finally sending the desired speed back to the vehicle where the actuation is carried out. 

\subsubsection{Remote Management} 
The ability to remotely manage ADAS systems enables the development, testing, and operation of these systems from a different geographical location than the physical twin.
In addition to displaying real-time state data, the UI will also enable the management and operation of ACC remotely. This includes enabling/disabling ACC, setting the desired speed, and sending an emergency braking command. 

To implement these services, the DT must support low end-to-end latency communication with the vehicle, contain data collection and reporting capabilities, and more---all while respecting the safety-critical nature of the system.
For convenience and following community best practices, we refer to the collection of the DT Entity, physical and middleware components, and the UI as the digital twin system (DTS). 
The users of our system can be grouped into two classes: operators and developers. Operators use the DT to monitor and remotely manage ACC, and developers use the DT to troubleshoot during development using the data collection and monitoring capabilities.
In the \textit{Liberty to act vs. Ability to act} taxonomy from \textcite{david2024infonomics}, this DTS falls under the Fully Autonomous Digital Twin, with the Human-in-the-Loop.

\subsection{Physical Twin}

The physical twin (PT) is a 1/10\textsuperscript{th}-scale autonomous vehicle (\figref{fig:pt}), built off a Traxxas teleoperated vehicle\footnote{\url{https://traxxas.com/}} by retrofitting it with autonomy components. In the retrofitting, we follow the process provided by the open-source RoboRacer project~\cite{roboracer}.
As shown in Fig.~\ref{fig:hw-arch}, the physical architecture consists of three layers.
The \textit{lower level chassis} is responsible for controlling the steering and the speed of the vehicle. It is a stripped-down version of the off-the-shelf Traxxas Slash 4x4 Premium vehicle.
The \textit{upper level chassis} is mounted on top of the lower level chassis to provide a physical platform for mounting the technology required for enabling autonomous control and digital twinning. The list of these autonomy components include an \textit{NVIDIA Jetson NX antenna} and \textit{NVIDIA Jetson Orin Nano} microcomputer, and onboard sensors including a \textit{Hokuyo UST-10LX} LiDAR, an \textit{Intel Deepsense D435i} Depth Camera, and a \textit{Vedder Electronic Speed Controller}.

We use \textit{NVIDIA Jetpack 5.0} as the Software Development Kit (SDK) for programming the Jetson Orin Nano, and \textit{Robot Operating System 2} (ROS2)~\cite{macenski2022robot} as the middleware to integrate application modules. Application modules are implemented in Python 3.8 (rapid prototyping), C++ (ROS2 node lifecycle management), and MATLAB/Simulink (for more complex systems, e.g., point cloud clustering and tracking).
\section{Engineering a DT for Adaptive Cruise Control}\label{sec:case}

We now provide a detailed exposition of the case study, starting from the definition of DTS requirements (\secref{sec:requirements}). We then elaborate on the architectural design (\secref{sec:architecture}) and the implementation of DTS services (\secref{sec:services}).

\subsection{DTS Requirements}\label{sec:requirements}
To drive its development, we first define system requirements for the DTS.
We rely on a case-based generalization~\cite{wieringa2015six} from real projects -- an approach that is particularly effective when constructing middle-range theories that balance generality with practicality, such as the ones in engineering sciences. We draw on our expertise with automotive systems and our ongoing projects in the area as we define the following representative requirements for the DT.

\begin{table*}[t]
    \centering
    \caption{SW Components and corresponding FEs}
    \label{tab:swc-to-fe}
    \begin{tabular}{@{}lp{10cm}l@{}}
    
    \toprule
    \multicolumn{1}{c}{\textbf{SW Component}}
    & \multicolumn{1}{c}{\textbf{Description}}
    & \multicolumn{1}{c}{\textbf{ISO 23247 FE}}\\

    \midrule

    \multicolumn{3}{c}{\textbf{Physical Twin Entity}}\\

    Point Cloud Clustering \& Tracking & Clusters raw LiDAR point clouds via DBSCAN, then tracks using a Kalman filter. & Data Pre-Processing FE\\

    Actuation & Converts desired speed into motor voltage commands. & Actuation FE\\

    \midrule
    
    \multicolumn{3}{c}{\textbf{Digital Twin Entity}}\\

    Visualization Pre-Processing & Converts data into a format that the 3D Visualization can process. & Presentation FE\\

    Data Collection & Collects object tracks and ego vehicle state. & Data Collecting FE\\

    Report Generation & Generates human-readable reports of object tracks and vehicle state as a timeseries. & Reporting FE\\

    Data Storage & Gathers object tracks and ego vehicle state into a SQL database. & \textit{N/A}\\

    ACC & Determines lead vehicle and computes following speed. & Controlling FE \\

    \midrule
    
    \multicolumn{3}{c}{\textbf{User Entity}}\\

    3D Visualization & Presents 3D object tracks, vehicle state information, and projected driving path. & User Interface FE\\

    Control Panel & Interface for the user to send remote management commands. & User Interface FE\\

    \midrule

    \multicolumn{3}{c}{\textbf{Cross-System Entity}}\\
    
    Data Translation & Translation of exchanged data between entities and FEs. & Data Translation FE \\
    
    \bottomrule    
    \end{tabular}
\end{table*}

\begin{enumerate}[topsep=0.5em, itemsep=0.5em, label=\textbf{R{\arabic*}.}]
    \item \textbf{The DTS shall use a standardized communication interface between all components of the system.} \label{rInternalComm}
    Ensures alignment with good system design principles, such as modularity and loose coupling, to hide internal complexities of each component.

    \item \textbf{The DTS shall collect real-time data about the ego vehicle's state.} \label{rCollect}
    The DTS shall collect the following ACC-relevant data in real-time:
    \begin{itemize}[noitemsep,topsep=0pt]
        \item ego vehicle's speed;
        \item ego vehicle's steering angle;
        \item detected objects in the vehicle's proximity.
    \end{itemize}

    \item \textbf{The DTS shall store the ego vehicle state data on a device external to the vehicle.} \label{rDataStorage}
    Due to large quantities of real-time data and limited storage on the vehicle, data must be stored externally to enable data monitoring.
    
    \item \textbf{The DTS shall report real-time ego vehicle information through a UI.} \label{rUIMonitor}
    Drivers are more likely to trust ADAS systems with transparent human-machine interfaces that report relevant information of the ADAS decision-making process~\cite{liu2022trust}.
    Thus, the UI shall display the following information about the ego vehicle in real-time:
    \begin{itemize}[noitemsep,topsep=0pt]
        \item ego vehicle's speed;
        \item ego vehicle's steering angle;
        \item tracks surrounding the ego vehicle;
        \item projected driving path of the ego vehicle.
    \end{itemize}

    \item \textbf{The DTS shall include a user interface to remotely manage the ACC sub-system.} \label{rUIControl}
    To enable remote management, the UI implements the following commands:
    \begin{itemize}[noitemsep,topsep=0pt]
        \item enable and disable ACC;
        \item modify the desired speed of ACC;
        \item perform emergency braking of the ego vehicle.
    \end{itemize}

    \item \textbf{The DTS shall actuate the ego vehicle in response to user commands and ACC control.} \label{rControl}
    The system shall implement ACC, as outlined in ISO 15622~\cite{iso15622}. The basic control strategy and functionality, described in \cite[Sec 6.1--6.2]{iso15622} shall be implemented for performance class I ACC (no road curvature capabilities), while the remaining requirements are not considered due to ACC being implemented on a 1/10\textsuperscript{th}-scale vehicle. 

    \item \textbf{The DTS shall maintain network latency of at most 100 ms from the vehicle to the DT Entity.} \label{rLatency}
    To enable computation offloading for a real-time system, low end-to-end latency is a must.
    This figure comes from the 3rd Generation Partnership Plan (3GPP) Technical Specification, 22.185~\cite{etsi-ts22185}, a specification of service requirements for Vehicle-to-Vehicle (V2V) communication.

    \item \textbf{Components within the DTS shall have clocks synchronized to an accuracy of 100 ms.} \label{rTimeSync}
    Clock synchronization is especially critical in distributed systems for the purposes of event ordering, data integrity, and coordination between devices. To enable computation offloading, devices must share the same notion of time.
    Many standards mandate stricter clock synchronization   than 100 ms for safety-critical vehicular applications. We require 100 ms accuracy due to hardware limitations. 
\end{enumerate}

\begin{figure*}
    \centering
    \includegraphics[width=0.88\linewidth]{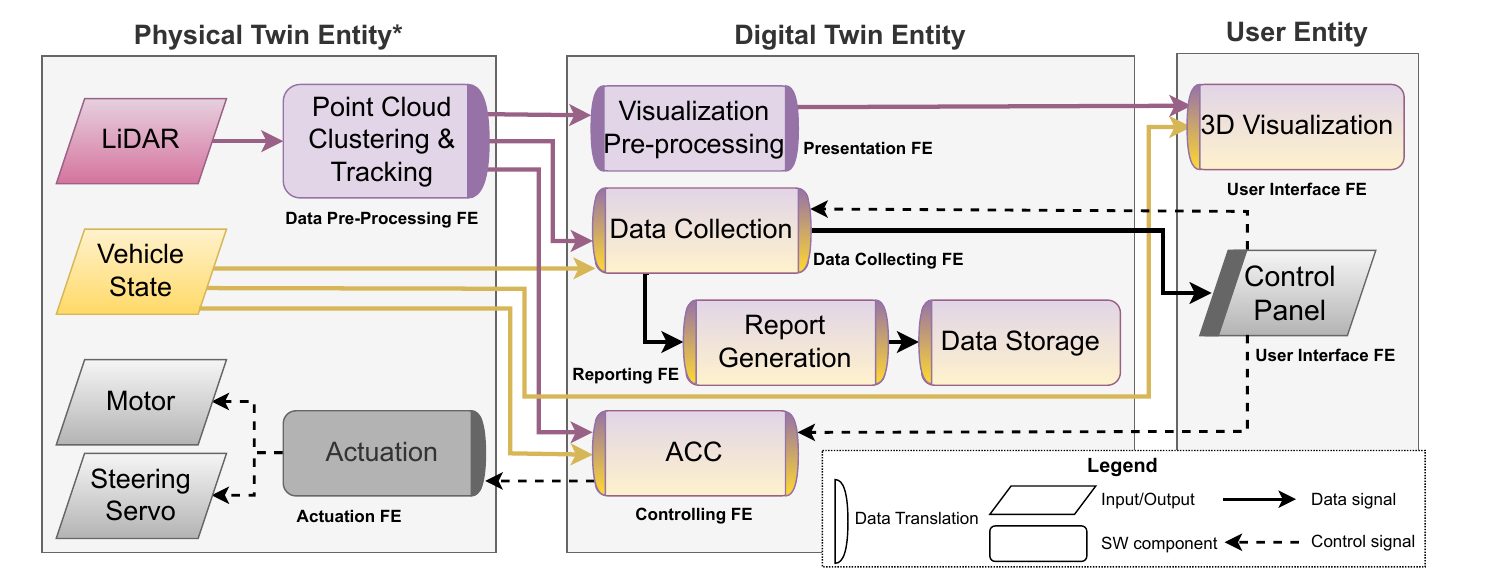}
    \caption{Architecture and deployment view of the DTS. Colors show how sensor data propagates in the system.}
    \label{fig:architecture}
    \vspace{-0.5em}
    \caption*{$^\ast$The Physical Twin Entity is not defined in ISO 23247. Closest to it is the Observable Manufacturing Element (OME).}
\end{figure*}

\subsection{Architecture}\label{sec:architecture}

We draw on the ISO 23247 RA in the architectural design phase. First, we identify key software (SW) components (\tabref{tab:swc-to-fe}) and their relationships to each other, and to the sensors and actuators of the physical twin (\figref{fig:architecture}).

Next, we map the SW components onto FEs and implement them. To enable their execution on the vehicle, we implement each FE as an individual ROS2 package. Each package contains one or more ROS2 nodes implementing the functionality provided by the FE.
This mapping helps organize the SW components by the entity-based structure of ISO 23247.

User-facing components are organized into the \textit{User Entity}, within which the singular \textit{User Interface FE} is instantiated in the \textit{3D Visualization} and \textit{Control Panel} components.
SW components that implement data management and control mechanisms are organized into the \textit{Digital Twin Entity}. 
Data pre-processing, actuation, and parts of the device communication functionality are organized into the \textit{Physical Twin Entity}.
The existence of this entity marks a departure from the ISO 23247 RA. Most similar to the Physical Twin Entity in the standard is the OME, which hosts resource-specific functionality.
By contrast, in our design, the Physical Twin Entity reflects the inherent resources of the twinned system (e.g., compute, sensors, actuators) and can host general automotive functionality, such as \textit{Point Cloud Clustering \& Tracking}.
Considering that each entity has been deployed on a different hardware component, such deviations from the standard are substantial as they may impact the ownership of the functionality and the engineering process of the DTS.

\begin{table}[t]
  \centering
  \caption{SW Components and corresponding DTS Services}
  \label{tab:SWC_SVC}
  \begin{tabular}{@{} 
      l      
      c      
      c      
      c      
    @{}}
    \toprule
    \textbf{SW Component}
      & \textbf{\begin{tabular}[c]{@{}c@{}}Data\\Monitoring\end{tabular}}
      & \textbf{\begin{tabular}[c]{@{}c@{}}Computation\\Offloading\end{tabular}}
      & \textbf{\begin{tabular}[c]{@{}c@{}}Remote\\Management\end{tabular}}
    \\
    
    \midrule

    \multicolumn{4}{c}{\textbf{Physical Twin Entity}}\\
      
    \begin{tabular}[t]{@{}l@{}}Point Cloud \\Clustering \& Tracking\end{tabular}
      & $\checkmark$ & $\checkmark$ & $\checkmark$\\

    Actuation
      &$\textendash$&$\textendash$& $\checkmark$ \\

    \midrule
    
    \multicolumn{4}{c}{\textbf{Digital Twin Entity}}\\
    
    Visualization Pre-Processing
      & $\checkmark$ &$\textendash$&$\checkmark$\\

    Data Collection
      & $\checkmark$ & $\textendash$ & $\textendash$ \\
      
    Report Generation
      & $\checkmark$ &$\textendash$ &$\textendash$\\

    Data Storage
      & $\checkmark$ &$\textendash$ &$\textendash$\\

    ACC
      &$\textendash$& $\checkmark$ &$\checkmark$\\

    \midrule
    
    \multicolumn{4}{c}{\textbf{User Entity}}\\
    
    3D Visualization
      & $\checkmark$ &$\textendash$ & $\checkmark$ \\

      Control Panel
      & $\textendash$ &$\textendash$ & $\checkmark$ \\

    \midrule

    \multicolumn{4}{c}{\textbf{Cross-System Entity}}\\
    
    Data Translation
      & $\checkmark$ & $\checkmark$ & $\checkmark$ \\
      
    \bottomrule
  \end{tabular}
\end{table}

Not every ISO 23247 FE was required for our DTS. At the same time, the \textit{User Interface FE} has been split into two sub-components: visualization and control panel. Finally, one SW component, \textit{Data Storage} did not have a naturally aligned FE.


\begin{figure*}[t]
    \begin{subfigure}{.4\textwidth}
        \centering
        \includegraphics[width=0.85\textwidth]{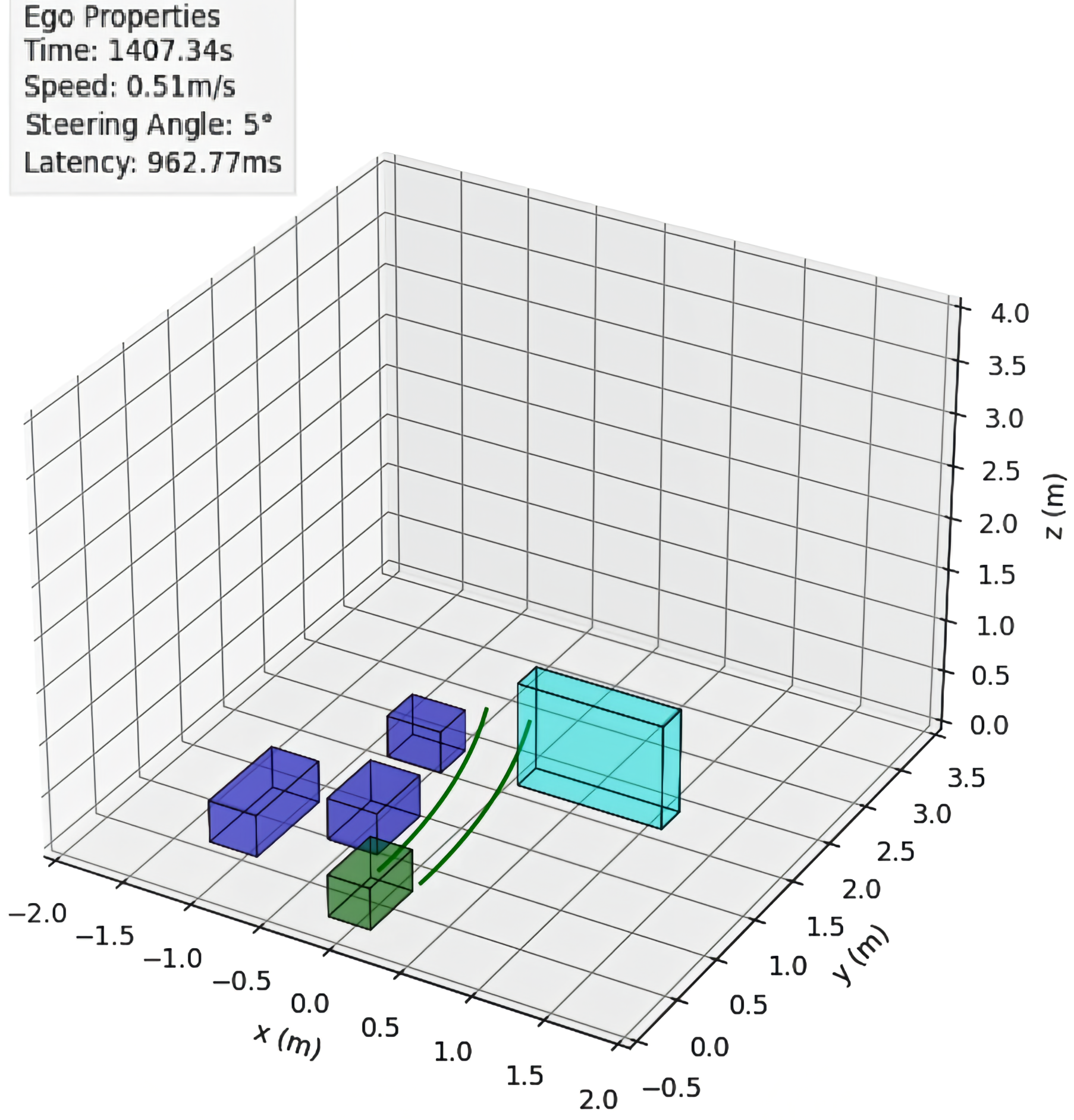}
        \caption{3D Visualization of the vehicle and its environment}
        \label{fig:cuboidVisualizer}
    \end{subfigure}
    \hfill
    \begin{subfigure}{.6\textwidth}
        \centering
        \includegraphics[width=.95\linewidth]{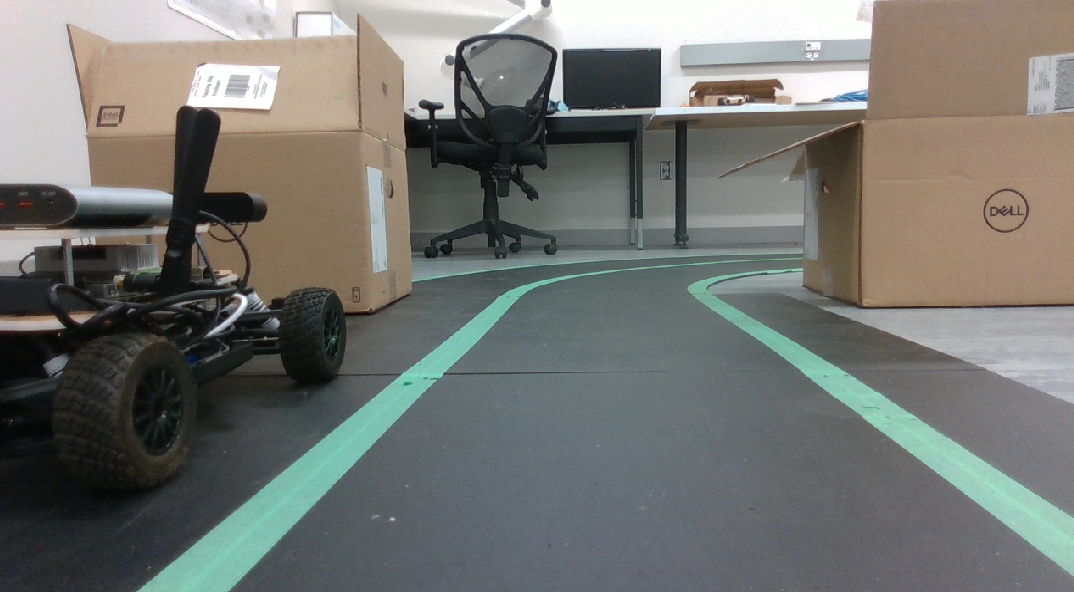}
        \caption{Corresponding view from the on-board camera}
        \label{fig:cameraView}
    \end{subfigure}
    
    \caption{Visualizations available from the virtual cockpit}
    
    \label{fig:ui}
\end{figure*}

\subsection{Implementation of Services}\label{sec:services}

Below, we describe each DTS service as a composition of multiple SW components, summarized in \tabref{tab:SWC_SVC}.

All of the DTS services rely on the Data Translation component, which corresponds to the ISO 23247 Data Translation FE. This component serializes data, aligns message types, and performs unit conversions for every data stream in the system.
Communication between SW components takes place within a ROS2 network, while inter-entity communication occurs over a Wi-Fi network using UDP in the transport layer.

\subsubsection{Data Monitoring}
To enable data monitoring, ACC-relevant data such as vehicle state information, object tracks, and projected driving path are displayed in real-time on the 3D Visualization, as shown in Fig.~\ref{fig:cuboidVisualizer}. Raw point cloud data from the LiDAR sensor must be processed by the \textit{Point Cloud Clustering \& Tracking} SW component, which turns the raw point cloud, represented as individual points on a 2D plane, into object tracks. 
The data then goes through the \textit{Visualization Pre-Processing}, which transforms object track data into a 3D representation. Subsequently, this information is visualized by the \textit{3D Visualization} SW component.
Furthermore, the DTS contains functionality to save and generate reports of this data. After being requested by the user, processed sensor data and vehicle state data are compressed into a SQL database for storage by the \textit{Data Collection} SW component. Finally, a report of the data is generated by the \textit{Reporting} SW component in the form of a timeseries CSV file. 

\subsubsection{Computation Offloading}
After object tracks have been computed by the \textit{Point Cloud Clustering \& Tracking} SW component, these object tracks are then sent to the DT Entity, which hosts the \textit{ACC} SW component. The \textit{ACC} SW component determines which object track represents the lead vehicle and computes a following speed, which is a function of the distance to the lead vehicle. If no lead vehicle is present, \textit{ACC} requests the desired speed provided by the \textit{Control Panel} SW component.
The DT Entity then sends the desired speed to the \textit{Actuation} SW component, turning the desired speed into a motor command via a PID controller.

\subsubsection{Remote Management}
\begin{figure}[t]
    \centering
    \includegraphics[width=.9\linewidth]{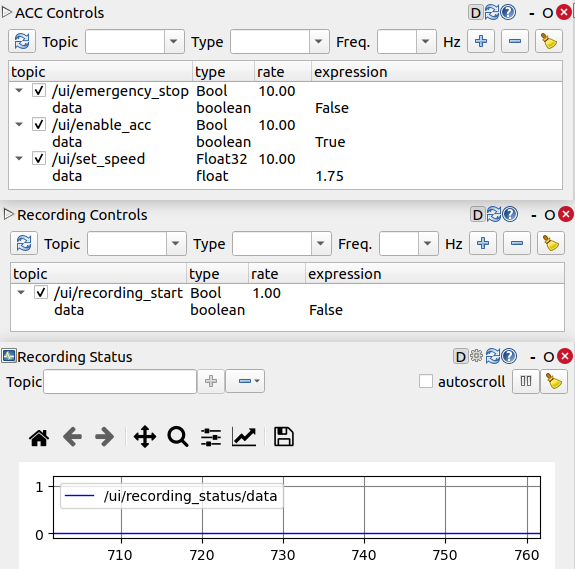}
    \caption{Control panel for remote management}
    \label{fig:controlPanel}
\end{figure}

A virtual cockpit~\cite{dalibor2020towards}
allows users to remotely manage the \textit{ACC} via the \textit{Control Panel}, as shown in \figref{fig:controlPanel}. Users can remotely enable or disable \textit{ACC}, input the desired speed, or send emergency braking commands. Additionally, the virtual cockpit contains a 3D Visualization of object tracks (\figref{fig:cuboidVisualizer}), with detected vehicles and larger obstacles represented by blue boxes and cyan boxes, respectively. The ego vehicle and the instantaneous driving path are represented by the green box and curves emitting from it. Real-time vehicle state information is displayed in the top left corner. Finally, users can choose to view the real-time camera stream from the forward-facing depth camera (\figref{fig:cameraView}).

\section{Reflection and Conclusions}\label{sec:conclusion}

We reported on a case study of an automotive DT, through which we investigate the fitness of the ISO 23247 standard in engineering such systems.
Our experiences highlight both benefits and limitations of the standard in automotive settings. 


On a positive note, we found the standard's reference architecture (RA) to be \textbf{a useful starting point for automotive DTs} as it provides guidance for organizing blocks of functionality.
Additionally, we found the standard \textbf{easy to use} as it does not assume substantial DT expertise and has a comfortable learning curve. Our case was developed by two graduate students, starting from a course project.\footnote{\url{https://academiccalendars.romcmaster.ca/preview_course_nopop.php?catoid=55&coid=282113}}
Through this case study, we also highlight the \textbf{convenient generality} of ISO 23247. The independence from specific format and protocol requirements, as well as the ability to rely on domain-specific technologies in a straightforward manner, facilitated the transfer of the RA to a new domain (automotive).

On a critical note, rigid adherence to the RA may lead to an \textbf{under-utilization of resources}.
In contrast to OMEs, vehicles are resource-rich systems of their own. Thus, we recommend promoting the physical twin to a first-class entity in automotive DT architectures. This will allow for deploying FEs on the vehicle, potentially in a flexible and dynamic fashion.
For example, our case study deploys the Data Pre-Processing FE and the Actuation FE onto the physical twin, as opposed to the DT Entity as prescribed by the ISO 23247 RA. Flexibility in the deployment of FEs is at the heart of existing automotive standards---e.g., AUTOSAR, the \textit{de facto} automotive software architecture standard---where SW components are defined in a hardware-agnostic manner, making it easier to retrofit existing automotive architectures into automotive DT architectures.
Our case study further suggests that the RA is \textbf{insufficient for safety-critical applications such as automotive}.
Therefore, we recommend safety to be promoted to a first-class citizen in automotive DT standards by relying on existing automotive safety standards---e.g., ISO 26262~\cite{iso26262}.

Our work is a first look into the benefits and challenges of adopting a standardized DT architecture for automotive DTs. In order to aid the development of proper domain-specific standards, we call for additional research into the topic, with a particular emphasis on safety, security, data management, and considering the regulated nature of the automotive domain, compatibility with existing standards. Additionally, we encourage the development of automotive DT case studies and exemplars to provide empirical evidence for further research.



In future work, we plan to research the safety aspects of automotive DTs in more detail and elicit recommendations for standardization bodies.
The prototype system used in the case study is maintained in our host institute, the McMaster Centre for Software Certification (McSCert),\footnote{\url{https://www.mcscert.ca/}} and used as an open-source experiment system for automotive DT research.

\clearpage

\printbibliography

\end{document}